\pdfoutput=1

\documentclass[11pt]{article}

\usepackage{EACL2023}

\usepackage{times}
\usepackage{latexsym}

\usepackage[T1]{fontenc}

\usepackage[utf8]{inputenc}

\usepackage{microtype}
\usepackage{adjustbox}
\usepackage{url}
\usepackage{amssymb}
\usepackage{amsmath}
\usepackage{multirow}
\usepackage{graphicx} 
\usepackage{graphics}
\usepackage{multirow}
\usepackage{float}
\usepackage{fancyvrb}
\usepackage{amssymb}
\usepackage{amsmath}
\usepackage{amsfonts}
\usepackage{amssymb}
\usepackage{array}
\usepackage{ragged2e}
\usepackage{tabu}
\usepackage{algorithmicx}
\usepackage{algorithm}
\usepackage{algpseudocode}

\usepackage{adjustbox}

\usepackage{inconsolata}
\usepackage{booktabs}

\title{Semantics-Consistent Cross-domain Summarization via \\Optimal Transport Alignment}

\author{Jielin Qiu$^{\diamondsuit}$, ~Jiacheng Zhu$^{\diamondsuit}$, ~Mengdi Xu$^{\diamondsuit}$, ~Franck Dernoncourt$^{\spadesuit}$, \\
\textbf{~Zhaowen Wang$^{\spadesuit}$,  ~Trung Bui$^{\spadesuit}$, ~Bo Li$^{\heartsuit}$, ~Ding Zhao$^{\diamondsuit}$, ~Hailin Jin$^{\spadesuit}$ }\\
  $^{\diamondsuit}$Carnegie Mellon University, 
  ~$^{\spadesuit}$Adobe Research, 
  ~$^{\heartsuit}$University of Illinois Urbana-Champaign \\
  \small{$\left\{\text{jielinq,jzhu4,mengdixu}\right\}$@andrew.cmu.edu, $\left\{\text{dernonco,zhawang,bui,hljin}\right\}$@adobe.com,
  lbo@illinois.edu}
  }

\begin{document}
\maketitle
\begin{abstract}
Multimedia summarization with multimodal output (MSMO) is a recently explored application in language grounding. It plays an essential role in real-world applications, i.e., automatically generating cover images and titles for news articles or providing introductions to online videos. However, existing methods extract features from the whole video and article and use fusion methods to select the representative one, thus usually ignoring the critical structure and varying semantics. In this work, we propose a Semantics-Consistent Cross-domain Summarization (SCCS) model based on optimal transport alignment with visual and textual segmentation. In specific, our method first decomposes both video and article into segments in order to capture the structural semantics, respectively. Then SCCS follows a cross-domain alignment objective with optimal transport distance, which leverages multimodal interaction to match and select the visual and textual summary. We evaluated our method on three recent multimodal datasets and demonstrated the effectiveness of our method in producing high-quality multimodal summaries.
\end{abstract}

\section{Introduction}
New multimedia content in the form of short videos and corresponding text articles has become a significant trend in influential digital media, including CNN, BBC, Daily Mail, social media, etc.
This popular media type has shown to be successful in drawing users' attention and delivering essential information in an efficient manner.

\begin{figure}[htp]
  \centering
  \includegraphics[width=0.99\linewidth]{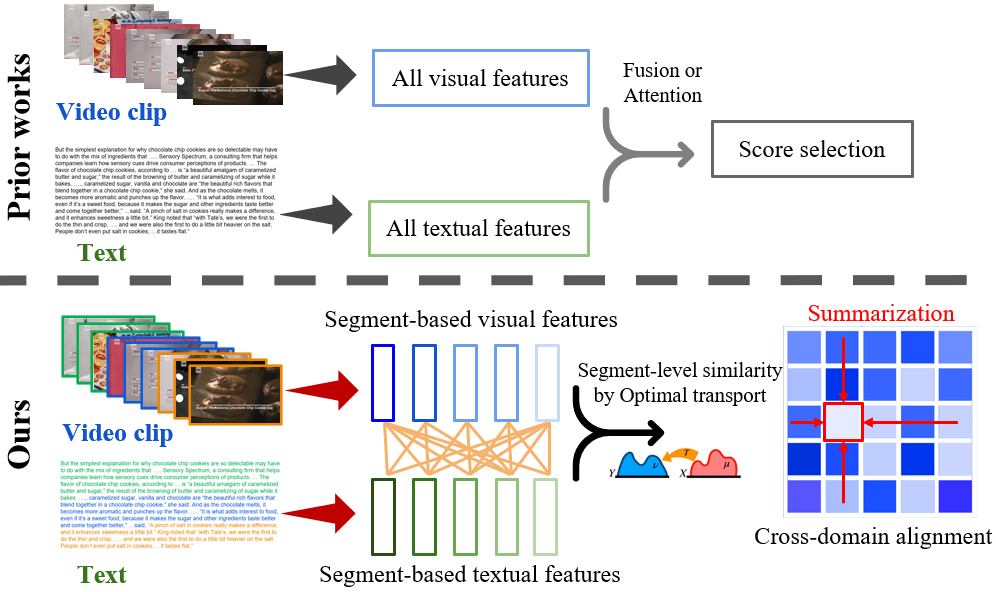}
  \caption{We proposed a segment-level cross-domain alignment model to preserve the structural semantics consistency within two domains for MSMO. We solve  an optimal transport problem to optimize the cross-domain distance, which in turn finds the optimal match.}
  \label{fig:intro}
\end{figure}

\begin{figure*}[htp]
  \centering
  \includegraphics[width=0.99\linewidth]{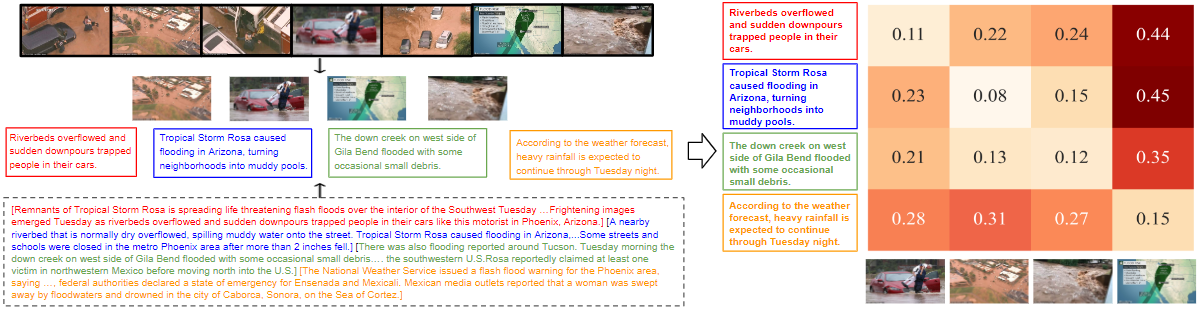}
  \vspace{-8pt}
  \caption{An illustration of the summarization process given by our SCCS method. Here we conduct OT-based cross-domain alignment to each keyframe-sentence pair and a smaller OT distance means better alignment. (For example, the best-aligned text and image summary ($0.08$) delivers the flooding content clearly and comprehensively.)}
  \label{fig:illu}
\end{figure*}

Multimedia summarization with multimodal output (MSMO) has recently drawn increasing attention. Different from traditional video or textual summarization \citep{Gygli2014CreatingSF,Jadon2020UnsupervisedVS}, where the generated summary is either a keyframe or textual description, MSMO aims at producing both visual and textual summaries simultaneously, making this task more complicated. 
Previous works addressed the MSMO task by processing the whole video and article together and used fusion or attention methods to generate scores for summary selection, which overlooked the structure and semantics of different domains \citep{Duan2022MultimodalAU,Haopeng2022VideoSB,Sah2017SemanticTS,Zhu2018MSMOMS,Mingzhe2020VMSMOLT,Fu2021MMAVSAF,Fu2020MultimodalSF}. 
However, we believe the structure of semantics is a crucial characteristic that can not be ignored for multimodal summarization tasks. 
Based on this hypothesis, we proposed Semantics-Consistent Cross-domain Summarization (SCCS) model, which explores segment-level cross-domain representations through Optimal Transport (OT) based multimodal alignment to generate both visual and textual summaries. 


The comparison of our approach and previous works is illustrated in Figure~\ref{fig:intro}. We regard the video and article as being composed of several topics related to the main idea, while each topic specifically corresponds to one sub-idea. 
Thus, treating the whole video or article uniformly and learning a general representation 
will ignore these structural semantics and leads to biased summarization. 
To address this problem, instead of learning averaged representations for the whole video \& article, we focus on exploiting the original underlying structure. 
Our model first decomposes the video \& article into segments to discover the content structure, then explores the cross-domain semantics relationship at the segment level.
We believe this is a promising approach to exploit the \textit{consistency} lie in the structural semantics between different domains. 
Since MSMO generates both visual \& textual summaries, 
We believe the optimal summary comes from the video and text pair that are both 1) semantically consistent and 2) best matched globally in a cross-domain fashion.
In addition, our framework is more computationally efficient as it 
conducts cross-domain alignment on segment-level rather than taking the whole videos/articles as inputs. 

Our contributions can be summarized as follow: 
\vspace{-20pt}
\begin{itemize}
    \item We propose SCCS (Semantics-Consistent Cross-domain Summarization), a segment-level alignment model for MSMO tasks. 
    \vspace{-10pt}
    \item Our method preserves the structural semantics and explores the cross-domain relationship through optimal transport to match and select the visual and textual summary.
    \vspace{-8pt}
    \item Our method serves as a hierarchical MSMO framework that provides better interpretability via Optimal Transport alignment. 
    \vspace{-8pt}
    \item We provide both qualitative and quantitative results on three public datasets. Our method outperforms baselines and provides good interpretation of learned representations.
\end{itemize}

\begin{figure*}[htp]
  \centering
  \includegraphics[width=0.99\linewidth]{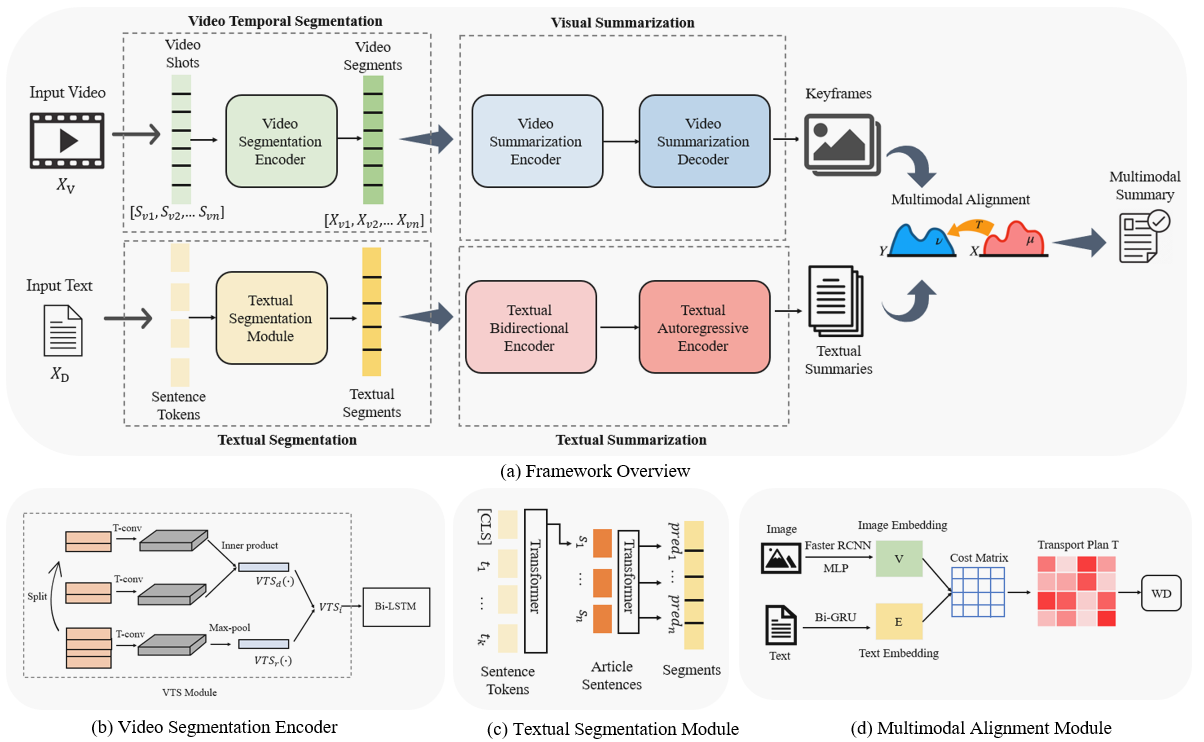}
  \vspace{-8pt}
  \caption{(a) The computational framework of our SCCS model, which takes a multimedia input (video+text) and generates multimodal summaries. The framework includes five modules for: video temporal segmentation, visual summarization, textual segmentation, textual summarization, and multimodal alignment. (b) The structure of the video segmentation encoder. (c) The architecture of the textual segmentation module. (d) The multimodal alignment module for multimodal summaries. }
  \label{fig:SCCS}
\end{figure*}\section{Related Work}
\paragraph{\textbf{Optimal Transport}}
Optimal Transport (OT) studies the geometry of probability spaces \citep{Villani2003TopicsIO}, a formalism for finding and quantifying mass movement from one probability distribution to another. OT defines the Wasserstein metric between probability distributions, revealing a canonical geometric structure with rich properties to be exploited. The earliest contribution to OT originated from Monge in the eighteenth century. Kantorovich rediscovered it under a different formalism, namely the Linear Programming formulation of OT. With the development of scalable solvers, OT is widely applied to many real-world problems \citep{Flamary2021POTPO,Chen2020GraphOT,Yuan2020AdvancingWS,Klicpera2021ScalableOT,Alqahtani2021UsingOT,Lee2019HierarchicalOT,Chen2019ImprovingSL,Qiu2022OptimalTB,Duan2022MultimodalAU,Han2022AnEE,Zhu2022GeoECGDA}.

\paragraph{\textbf{Multimodal Alignment}}
Aligning representations from different modalities is an important
technique in multimodal learning. 
With the recent advancement, exploring the explicit relationship across vision and language has drawn significant attention \cite{Wang2020AnEF}. 
\citet{Torabi2016LearningLE,Yu2017EndtoEndCW} adopted attention mechanisms, \citet{Dong2021DualEF} composed pairwise joint representation, \citet{Chen2020FineGrainedVR,Wray2019FineGrainedAR,Zhang2018CrossModalAH} learned fine-grained or hierarchical alignment,  \citet{Lee2018StackedCA,Wu2019UnifiedVE} decomposed the inputs into sub-tokens, \citet{Velickovic2018GraphAN,Yao2018ExploringVR} adopted graph attention for reasoning, and \citet{Yang2021TACoTC} applied contrastive learning algorithms.

\paragraph{\textbf{Multimodal Summarization}}
Multimodal summarization explored multiple modalities, i.e., audio signals, video captions, Automatic Speech Recognition (ASR) transcripts, video titles, etc, for summary generation. \citet{Otani2016VideoSU,Yuan2019VideoSB,Wei2018VideoSV,Fu2020MultimodalSF} learned the relevance or mapping in the latent space between different modalities.
In addition to only generating visual summaries, \citet{Li2017MultimodalSF,Atri2021SeeHR,Zhu2018MSMOMS} generated textual summaries by taking audio, transcripts, or documents as input along with videos or images, using seq2seq model \citep{Sutskever2014SequenceTS} or attention mechanism \citep{Bahdanau2015NeuralMT}. Recent trending on the MSMO task have also drawn much attention \citep{Zhu2018MSMOMS,Mingzhe2020VMSMOLT,Fu2021MMAVSAF,Fu2020MultimodalSF,Zhang2022HierarchicalCS}.

\section{Methods}
Our SCCS is a segment-level cross-domain semantics alignment model for the MSMO task, where MSMO aims at generating both visual and language summaries. We follow the problem setting in \citet{Mingzhe2020VMSMOLT}, for a multimedia source with documents/articles and videos, the document $X_D=\{x_1, x_2, ..., x_{d} \}$ has $d$ words, and the ground truth textual summary $Y_D = \{ y_1, y_2, ..., y_{g} \}$ has $g$ words. The corresponding video $X_V$ is aligned with the document, and there exists a ground truth cover picture $Y_V$ that can represent the most important information to describe the video. Our SCCS model generates both textual summary $Y_D'$ and video keyframe $Y_V'$.

Our SCCS model consists of five modules, as shown in Figure~\ref{fig:SCCS}(a): video temporal segmentation (Section~\ref{sec:video_temp}), visual summarization (Section~\ref{sec:visual_sum}), textual segmentation (Section~\ref{sec:text_seg}), textual summarization (Section~\ref{sec:text_sum}), and cross-domain alignment (Section~\ref{sec:align}).
Each module will be introduced in the following subsections.

\subsection{Video Temporal Segmentation}\label{sec:video_temp}
Video temporal segmentation aims at splitting the original video into small segments, which the summarization tasks build upon.
VTS is formulated as a binary classification problem on the segment  boundaries, similar to \citet{Rao2020ALA}. For a video $X_V$, the video segmentation encoder separates the video sequence into segments $[X_{v1},X_{v2}, ..., X_{vm}]$, where $n$ is the number of segments. 

As shown in Figure~\ref{fig:SCCS}(b), the video segmentation encoder  contains a VTS module and a Bi-LSTM. Video $X_V$ is first split into shots $[S_{v1},S_{v2},...,S_{vn}]$ \citep{PySceneDetect}, then the VTS module takes a clip of the video with $2 \omega_b$ shots as input and outputs a boundary representation $b_i$. The boundary representation captures both differences and relations between the shots before and after. VTS consists of two branches, VTS$_d$ and VTS$_r$, as shown in Equation~\ref{Equation:VTS_i}. 
\begin{equation}\scriptsize
\begin{aligned}
&  b_{i} 
={\text{VTS}}\left(\left[{S}_{vi-\left(\omega_{b}-1\right)}, \cdots, {S}_{vi+\omega_{b}}\right]\right)  \\
&=\left[\begin{array}{l}
\text{ VTS}_{d}\left(\left[{S}_{vi-\left(\omega_{b}-1\right)}, \cdots, \text{P}_{vi}\right],\left[{S}_{v(i+1)}, \cdots, {S}_{vi+\omega_{b}}\right]\right) \\
\text{ VTS}_{r}\left(\left[{S}_{vi-\left(\omega_{b}-1\right)}, \cdots, {P}_{vi}, {S}_{v(i+1)}, \cdots, {S}_{vi+\omega_{b}}\right]\right)
\end{array}\right]
\end{aligned} 
\label{Equation:VTS_i}
\end{equation}
VTS$_d$ is modeled by two temporal convolution layers, each of which embeds the $w_b$ shots before and after the boundary, respectively, following an inner product operation to calculate the differences. VTS$_r$ contains a temporal convolution layer followed a max pooling, aiming at capturing the relations of the shots. It predicts a sequence binary labels $[p_{v1},p_{v2},...,p_{vn}]$ based on the sequence of representatives $\text [b_1, b_2, ..., b_n]$. A Bi-LSTM \cite{Graves2005FramewisePC} is used with stride $\omega_t / 2$ shots to predict a sequence of coarse score $[s_1,s_2,...,s_n]$, as shown in Equation~\ref{Equation:prb}, 
\vspace{-8pt}
\begin{equation}\small
[s_{1},s_{2},...,s_{n}]=\text{Bi-LSTM}\left(\left[{b}_{1}, b_2, \cdots, {b}_{n}\right]\right)
\label{Equation:prb}
\end{equation}
where $s_i \in [0,1]$ is the probability of a shot boundary to be a scene boundary. The coarse prediction $\hat{p}_{vi} \in \{0, 1\}$ indicates whether the $i$-th shot boundary is a scene boundary by binarizing $s_i$ with a threshold $\tau$, $\small{\hat{p}_{vi}= \begin{cases}1 & \text { if } s_{i}>\tau \\ 0 & \text { otherwise }\end{cases} }$. The results with ${\hat{p}_{vi}=1}$ result in the learned video segments $[X_{v1},X_{v2}, ..., X_{vm}]$

\subsection{Textual Segmentation}\label{sec:text_seg}
The textual segmentation module takes the whole document or articles as input and splits the original input into segments based on context understanding. 
We used a hierarchical BERT as the textual segmentation module \citep{lukasik-etal-2020-text}. 
As shown in Figure~\ref{fig:SCCS}(c), the textual segmentation module contains two-level transformer encoders, where the first-level encoder is for sentence-level encoding, and the second-level encoder is for the article-level encoding. The hierarchical BERT starts by encoding each sentence with $\text{BERT}_{\text{LARGE}}$ independently, then the tensors produced for each sentence are fed into another transformer encoder to capture 
the representation of the sequence of sentences. All the sequences start with a [CLS] token to encode each sentence with BERT at the first level. If the segmentation decision is made at the sentence level, we use the [CLS] token as input of the second-level encoder. The [CLS] token representations from sentences are passed into the article encoder, which can relate the different sentences through cross-attention.

\subsection{Visual Summarization}\label{sec:visual_sum}
The visual summarization module generates visual keyframes from each segment as its corresponding summary. 
We use a encoder-decoder architecture with attention as the visual summarization module \citep{Ji2020VideoSW}, taking each video segment as input and outputting a sequence of keyframes. The encoder is a Bi-LSTM \citep{Graves2005FramewisePC} to model the temporal relationship of video frames, where the input is $X=[x_1, x_2, ..., x_m]$ and the encoding representation is $E =[e_1, e_2, ... e_m]$. The decoder is a LSTM \citep{Hochreiter1997LongSM} to generate output sequences $D=[d_1, d_2, ..., d_m]$. To exploit the temporal ordering across the entire video, we introduce attention mechanism:
$\mathrm{E}_{t}=\sum_{i=1}^{m} \alpha_{t}^{i} e_{i}, \text { s.t. } \sum_{i=1}^{n} \alpha_{t}^{i}=1 $, and
\vspace{-5pt}
\begin{equation}\small
\begin{gathered}
{\left[\begin{array}{c}
p\left(d_{t} \mid\left\{d_{i} \mid i<t\right\}, E_{t}\right) \\
s_{t}
\end{array}\right]=\psi\left(s_{t-1}, d_{t-1}, E_{t}\right)}
\end{gathered}
\end{equation}
where $s_t$ is the hidden state, $E_t$ is the attention vector at time $t$, $\alpha_t^i $ is the attention weight between the inputs and the encoder vector, $\psi$ is the decoder function. 
To obtain $\alpha_t^i $, the relevance score $e_t^i$ is computed by 
$e_t^i =\text{score} (s_{t-1}, e_i)$,
where the $\text{score}$ function  decides the relationship between
the $i$-th visual features $e_i$ and the output scores at time $t$:
$\beta_t^i = e^T_i W_a s_{t-1}$, $\alpha^i_t = \exp(\beta_t^i) / \sum^m_{j=1} \exp(\beta_t^j)$.

\subsection{Textual Summarization}\label{sec:text_sum}
Language summarization can produce a concise and fluent summary which should preserve the critical information and overall meaning. 
Our textual summarization module takes Bidirectional and Auto-Regressive Transformers (BART) \citep{Lewis2020BARTDS} as the summarization model to generate abstractive textual summary candidates. BART is a denoising autoencoder that maps a corrupted document to the original document it was derived from. 
As in Figure~\ref{fig:SCCS}(a), BART uses the standard sequence-to-sequence Transformer architecture, where both the encoder and the decoder include 12 layers. 
In addition to the stacking of encoders and decoders, cross attention between encoder and decoder is also applied. 
BART is trained by corrupting documents and then optimizing a reconstruction loss, where the pretraining task involves randomly shuffling the order of the original sentences and a novel in-filling scheme, where spans of text are replaced with a single mask token. 
BART works well for comprehension tasks, including achieving new state-of-the-art results on summarization tasks \citep{Lewis2020BARTDS}.

\subsection{Cross-Domain Alignment via Optimal Transport}\label{sec:align}
Cross-domain alignment module learns the alignment between keyframes and textual summaries to generate the final multimodal summaries. Our alignment module is based on OT, which has been explored in several cross-domain tasks  \cite{Chen2020GraphOT,Yuan2020AdvancingWS,Lu2021CrossdomainAR}. OT is the problem of transporting mass between two discrete distributions supported on latent feature space $\mathcal{X}$. Let $\boldsymbol{\mu}=\left\{\boldsymbol{x}_{i}, \boldsymbol{\mu}_{i}\right\}_{i=1}^{n}$ and $\boldsymbol{v}=\left\{\boldsymbol{y}_{j}, \boldsymbol{v}_{j}\right\}_{j=1}^{m}$ be the discrete distributions of interest, where $\boldsymbol{x}_{i}, \boldsymbol{y}_{j} \in \mathcal{X}$ denotes the spatial locations and $\mu_{i}, v_{j}$, respectively, denoting the non-negative masses. 
Without loss of generality, we assume $\small \sum_{i} \mu_{i}=\sum_{j} v_{j}=1$. $\pi \in \mathbb{R}_{+}^{n \times m}$ is a valid transport plan if its row and column marginals match $\mu$ and $\boldsymbol{v}$, respectively, which is $\sum_{i} \pi_{i j}=v_{j}$ and $\sum_{j} \pi_{i j}=\mu_{i}$. Intuitively, $\pi$ transports $\pi_{i j}$ units of mass at location $\boldsymbol{x}_{i}$ to new location $\boldsymbol{y}_{j}$. Such transport plans are not unique, and one often seeks a solution $\pi^{*} \in \Pi(\boldsymbol{\mu}, \boldsymbol{v})$ that is most preferable in other ways, where $\Pi(\boldsymbol{\mu}, \boldsymbol{v})$ denotes the set of all viable transport plans. OT finds a solution that is most cost effective w.r.t. cost function $C(\boldsymbol{x}, \boldsymbol{y})$:
$$\small 
\mathcal{D}(\boldsymbol{\mu}, \boldsymbol{v})=\sum_{i j} \pi_{i j}^{*} C\left(\boldsymbol{x}_{i}, \boldsymbol{y}_{j}\right)=\inf _{\pi \in \Pi(\mu, v)} \sum_{i j} \pi_{i j} C\left(\boldsymbol{x}_{i}, \boldsymbol{y}_{j}\right)
$$
where $\mathcal{D}(\boldsymbol{\mu}, \boldsymbol{v})$ is known as OT distance. $\mathcal{D}(\boldsymbol{\mu}, \boldsymbol{v})$ minimizes the transport cost from $\boldsymbol{\mu}$ to $\boldsymbol{v}$ w.r.t. $C(\boldsymbol{x}, \boldsymbol{y})$. When $C(\boldsymbol{x}, \boldsymbol{y})$ defines a distance metric on $\mathcal{X}$, and $\mathcal{D}(\boldsymbol{\mu}, \boldsymbol{v})$ induces a distance metric on the space of probability distributions supported on $\mathcal{X}$, it becomes the Wasserstein Distance (WD). 

The Cross-Domain alignment module is shown in Figure~\ref{fig:SCCS}(d), which is inspired by OT \citep{Yuan2020AdvancingWS}.  
The image features $V=\left\{\boldsymbol{v}_{k}\right\}_{k=1}^{K}$ are extracted from pre-trained ResNet-101 \citep{He2016DeepRL} concatenated to faster R-CNN \citep{Ren2015FasterRT} as \citet{Yuan2020AdvancingWS}. For text features, every word is embedded as a feature vector and processed by a  Bi-GRU to account for context \citep{Yuan2020AdvancingWS}. The extracted image and text embeddings are  $\mathbf{V}=\left\{\boldsymbol{v}_{i}\right\}_{1}^{K}$, $\mathbf{E}=\left\{\boldsymbol{e}_{i}\right\}_{1}^{M}$, respectively.

We take image and text sequence embeddings as two discrete distributions supported on the same feature representation space. Solving an OT transport plan between the two naturally constitutes a matching scheme to relate cross-domain entities \cite{Yuan2020AdvancingWS}. To evaluate the OT distance, we compute a pairwise similarity between $V$ and $E$ using cosine distance:
\vspace{-5pt}
\begin{equation}\small
\begin{aligned}
C_{km} = C(e_k, v_m) = 1-\frac{\boldsymbol{e}_{k}^{T} \boldsymbol{v}_{k}}{\left\|\boldsymbol{e}_{k}\right\|\left\|\boldsymbol{v}_{m}\right\|}
\end{aligned}
\end{equation}
\vspace{-5pt}
Then the OT can be formulated as:
\begin{equation}\small
\label{eq:ot_lp}
\mathcal{L}_{\mathrm{OT}}(\mathbf{V}, \mathbf{E})=\min _{\mathbf{T}} \sum_{k=1}^{K} \sum_{m=1}^{M} \mathbf{T}_{k m} \mathbf{C}_{k m}
\end{equation}
where $\small \sum_{m} \mathbf{T}_{k m}=\mu_{k}, \sum_{k} \mathbf{T}_{k m}=v_{m}, \forall k \in[1, K], m \in[1, M] ,$ ${\small \mathbf{T} \in \mathbb{R}_{+}^{K \times M}}$ is the transport matrix, $d_{k}$ and $d_{m}$ are the weight of $\boldsymbol{v}_{k}$ and $\boldsymbol{e}_{m}$ in a given image and text sequence, respectively. We assume the weight for different features to be uniform, i.e., $\mu_{k}=\frac{1}{K}, v_{m}=\frac{1}{M}$. 
The objective of optimal transport involves solving linear programming and may cause potential computational burdens since it has $O(n^3)$ efficiency. To solve this issue, we add an entropic regularization term equation (\ref{eq:ot_lp}) and the objective of our optimal transport distance becomes:
\vspace{-5pt}
\begin{equation}\small
    \mathcal{L}_{\mathrm{OT}}(\mathbf{V}, \mathbf{E})=\min _{\mathbf{T}} \sum_{k=1}^{K} \sum_{m=1}^{M} \mathbf{T}_{k m} \mathbf{C}_{k m} + \lambda H(\mathbf{T})
\end{equation}
where $H(\mathbf{T}) = \sum_{i,j} \mathbf{T}_{i,j} \log \mathbf{T}_{i,j}$ is the entropy, and $\lambda$ is the hyperparameter that balance the effect of the entropy term. Thus, we are able to apply the celebrated Sinkhorn algorithm \citep{cuturi2013sinkhorn} to efficiently solve the above equation in $O(n 
log n)$, where the algorithm is shown in Algorithm~\ref{alg:WD}.  The optimal transport distance computed via the Sinkhorn algorithm is differentiable and it can be implemented by \citet{Flamary2021POTPO}.
After training the alignment module,  the WD between each keyframe-sentence pair of all the visual \& textual summary candidates could be computed, where the best match is selected as the final multimodal summaries.

\begin{algorithm}[htp] \small
	\caption{Compute Alignment Distance}
	\begin{algorithmic}[1]
		\State \textbf{Input}: $ \mathbf{V}=\left\{\boldsymbol{v}_{i}\right\}_{1}^{K}, 
		\mathbf{E}=\left\{\boldsymbol{e}_{i}\right\}_{1}^{M},\beta$
		\State $\mathbf{C}=C(\mathbf{V}, \mathbf{E})$, $\sigma \leftarrow \frac{1}{m} \mathbf{1}_{m}, \mathbf{T}^{(1)} \leftarrow \mathbf{1} \mathbf{1}^{T}$
		\State $\mathbf{G}_{i j} \leftarrow \exp \left(-\frac{\boldsymbol{C}_{i j}}{\beta}\right)$
		\For{ t = 1,2,3,...,N}
		\State$\boldsymbol{Q} \leftarrow \boldsymbol{G} \odot \mathbf{T}^{(t)}$
		\For{l = 1,2,3,...,L}
		\State $\boldsymbol{\delta} \leftarrow \frac{1}{K \boldsymbol{Q} \sigma}, \boldsymbol{\sigma} \leftarrow \frac{1}{M \boldsymbol{Q}^{T} \boldsymbol{\delta}}$
		\EndFor
        \State $\mathbf{T}^{(t+1)} \leftarrow \operatorname{diag}(\boldsymbol{\delta}) \boldsymbol{Q} \operatorname{diag}(\boldsymbol{\sigma})$
        \EndFor
        \State $\mathbf{Dis} = <C^T, T>$
	\end{algorithmic}
	\label{alg:WD}
\end{algorithm}

\section{Datasets and Baselines}

\subsection{Datasets}\label{sec:dataset}
We evaluated our models on three datasets: VMSMO dataset,  Daily Mail dataset, and CNN dataset from \citet{Mingzhe2020VMSMOLT,Fu2021MMAVSAF,Fu2020MultimodalSF}. 
The VMSMO dataset contains  184,920 samples, including articles and corresponding videos. Each sample is assigned with a textual summary and a video with a cover picture. We adopted the available data samples from \citet{Mingzhe2020VMSMOLT}.
The Daily Mail dataset contains 1,970 samples, and the CNN dataset contains 203 samples, which include video titles, images, and captions, similar to \citet{Hermann2015TeachingMT}.
For data splitting, we take the same experimental setup as \citet{Mingzhe2020VMSMOLT} for the VMSMO dataset. For the Daily Mail dataset and CNN dataset, we split the data by 70\%, 10\%, 20\% for train, validation, and test sets, respectively, same as \citet{Fu2021MMAVSAF,Fu2020MultimodalSF}.

\subsection{Baselines}


We select state-of-the-art MSMO baselines and representative pure video summarization \& pure textual summarization baselines for comparison. For VMSMO dataset, we compare our method with (i) multimodal summarization baselines (MSMO, MOF, and DIMS \citep{Zhu2018MSMOMS,Zhu2020MultimodalSW,Mingzhe2020VMSMOLT}), (ii) video summarization baselines (Synergistic and PSAC \citep{Guo2019ImageQuestionAnswerSN,Li2019BeyondRP}),  and (iii) textual summarization baselines (Lead, TextRank, PG, Unified, and GPG \citep{Nallapati2017SummaRuNNerAR,Mihalcea2004TextRankBO,Abigail2017,Hsu2018AUM,Shen2019ImprovingLA}). 
For Daily Mail and CNN datasets, we comapre our method with (i) multimodal baselines (VistaNet, MM-ATG, Img+Trans, TFN,  HNNattTI, and M$^2$SM  \citep{vistanet,Zhu2018MSMOMS,hori2019end,zadeh-etal-2017-tensor,chen2018abstractive,Fu2021MMAVSAF,Fu2020MultimodalSF}), (ii) video summarization baselines (VSUMM, and DR-DSN  \citep{de2011vsumm,zhou2018deep}), and (iii) textual summarization baselines (Lead3, NN-SE, and M$^2$SM \citep{cheng2016neural,Fu2021MMAVSAF,Fu2020MultimodalSF}).

\begin{table*}[htp]\small
\centering
\caption{Comparison with multimodal baselines on the VMSMO dataset.}
\vspace{-8pt}
\begin{adjustbox}{width=0.8\linewidth}
\begin{tabular}{l|l|ccccccc} 
\toprule
\multirow{2}{*}{Category} &\multirow{2}{*}{Methods}   & \multicolumn{3}{c}{Textual} &\multicolumn{4}{c}{Video}   \\ 
\cmidrule(r){3-5}\cmidrule(r){6-9} &   &R-1 &R-2 &R-L & MAP &$R_{10}@1$ &$R_{10}@2$ &$R_{10}@5$ \\ 
\midrule
\multirow{2}{*}{Video summarization baselines} 
&Synergistic &-- &-- &-- & 0.558 &0.444 & 0.557 & 0.759 \\
&PSAC  &-- &-- &-- &0.524 &0.363 &0.481 &0.730    \\ 
\midrule
\multirow{6}{*}{Textual summarization baselines} 
&Lead    & 16.2  &5.3  &13.9    &-- &-- &-- &-- \\ 
&TextRank  & 13.7  & 4.0 & 12.5 &-- &-- &-- &--   \\   
&PG       & 19.4  & 6.8 & 17.4 &-- &-- &-- &--   \\
&Unified  & 23.0  & 6.0 & 20.9 &-- &-- &-- &--   \\
&GPG      & 20.1  & 4.5 & 17.3 &-- &-- &-- &--   \\
\midrule
\multirow{3}{*}{Multimodal baselines} 
&MSMO  & 20.1 &4.6 &17.3 &0.554 &0.361 &0.551 &0.820   \\
&MOF  &21.3 &5.7 &17.9 &0.615 &0.455 &0.615 &0.817    \\
&DIMS  &25.1 &9.6 &23.2 &0.654 &0.524 &0.634 &0.824 \\ 
\midrule
\multirow{3}{*}{Ours} 
&Ours-textual     &26.2   &9.6  &24.1  &--  &-- &-- &--   \\
&Ours-video      &--  & -- & --  &0.678  &0.561  &0.642  &0.863  \\ 
&Ours    &\textbf{27.1}   &\textbf{9.8}  &\textbf{25.4}     &\textbf{0.693}   &\textbf{0.582}   &\textbf{0.688}   &\textbf{0.895}   \\   
\bottomrule
\end{tabular}
\end{adjustbox}
\label{table:VMSMO_all}
\end{table*}

\section{Experiments}
\subsection{Experimental Setting and Implementation}
For the VTS module, we used the same model setting as \citet{Rao2020ALA,PySceneDetect} and same data splitting setting  as \citet{Mingzhe2020VMSMOLT,Fu2021MMAVSAF,Fu2020MultimodalSF} in the training process. 

The visual summarization model is pre-trained on the TVSum \citep{Song2015TVSumSW} and SumMe \citep{Gygli2014CreatingSF} datasets. TVSum dataset contains 50 edited videos downloaded from YouTube in 10 categories, and SumMe dataset consists of 25 raw videos recording various events. Frame-level importance scores for each video are provided for both datasets and used as ground-truth labels. 
The input visual features are extracted from pre-trained GoogLeNet on ImageNet, where the output of the pool5 layer is used as visual features.

For the textual segmentation module, due to the quadratic computational cost of transformers, we reduce the BERT’s inputs to 64 word-pieces per sentence and 128 sentences per document as \citet{lukasik-etal-2020-text}. We use 12 layers for both the sentence and the article encoders, for a total of 24 layers. In order to use the $\text{BERT}_{\text{BASE}}$ checkpoint, we use 12 attention heads and 768-dimensional word-piece embeddings. The hierarchical BERT model is pre-trained on the Wiki-727K dataset \cite{Koshorek2018TextSA}, which contains 727 thousand articles from a snapshot of the English Wikipedia. We used the same data splitting method as \citet{Koshorek2018TextSA}.  

For textual summarization, we adopted the pretrained BART model (bart-large-cnn\footnote{https://huggingface.co/facebook/bart-large-cnn}) from \citet{Lewis2020BARTDS}, which contains 1024 hidden layers and 406M parameters and has been fine-tuned using CNN and Daily Mail datasets. 

In cross-domain alignment module, the feature extraction and alignment module is pretrained by MS COCO dataset \citep{Lin2014MicrosoftCC} on the image-text matching task. We added the OT loss as a regularization term to the original matching loss to align the image and text more explicitly.

\subsection{Evaluation Metrics}
The quality of generated textual summary is evaluated by standard full-length Rouge F1 \citep{Lin2004ROUGEAP} following previous works \citep{Abigail2017,Chen2018IterativeDR,Mingzhe2020VMSMOLT}.  ROUGE-1 (R-1), ROUGE-2 (R-2), and ROUGE-L (R-L) refer to overlap of unigram, bigrams, and the longest common subsequence between the decoded summary and the reference, respectively \cite{Lin2004ROUGEAP}. 

For VMSMO dataset, the quality of chosen cover frame is evaluated by mean average precision (MAP) and recall at position $(R_n @ k)$ \citep{Zhou2018MultiTurnRS,Tao2019MultiRepresentationFN}, where $(R_n @ k)$ measures if the positive sample is ranked in the top $k$ positions of $n$ candidates. For Daily Mail  dataset and CNN dataset, we calculate the cosine image similarity (Cos) between image references and
the extracted frames from videos \citep{Fu2021MMAVSAF,Fu2020MultimodalSF}.

\subsection{Results and Discussion}
The comparison results on the VMSMO dataset of multimodal, video, and textual summarization are shown in Table~\ref{table:VMSMO_all}. 
Synergistic \cite{Guo2019ImageQuestionAnswerSN} and  PSAC \cite{Li2019BeyondRP} are pure video summarization approaches, which did not perform as good as multimodal methods, like MOF \cite{Zhu2020MultimodalSW} or DIMS \cite{Mingzhe2020VMSMOLT}, which means taking additional modality into consideration actually helps to improve the quality of the generated video summaries. 
Our method shows the ability to preserve the structural semantics and is able to learn the alignment between keyframes and textual deceptions, which shows better performance than the previous ones.
If comparing the quality of generated textual summaries, our method still outperforms the other multimodal baselines, like MSMO \cite{Zhu2018MSMOMS}, MOF \cite{Zhu2020MultimodalSW}, DIMS \cite{Mingzhe2020VMSMOLT}, and also traditional textual summarization methods, like Lead \cite{Nallapati2017SummaRuNNerAR}, TextRank \cite{Mihalcea2004TextRankBO}, PG  \cite{Abigail2017}, Unified \cite{Hsu2018AUM},  and GPG   \cite{Shen2019ImprovingLA}, 
showing the alignment  obtained by optimal transport can help to identify the cross-domain inter-relationships.

In Table~\ref{table:Dailymail_CNN_all}, we show the comparison results with multimodal baselines on the Daily Mail  and CNN datasets. 
We can see that for the CNN datasets, our method shows competitive results with
Img+Trans \cite{hori2019end}, TFN \cite{zadeh-etal-2017-tensor}, HNNattTI \cite{chen2018abstractive} and $\rm M^{2}$SM \cite{Fu2021MMAVSAF} on the quality of generated textual summaries. 
While on the Daily Mail dataset, our approach showed better performance on both textual summaries and visual summaries. 
We also compare with the traditional pure video summarization baselines \cite{de2011vsumm,zhou2018deep,Fu2021MMAVSAF} and pure textual summarization baselines \cite{Nallapati2017SummaRuNNerAR,cheng2016neural} on the Daily Mail dataset, and the results are shown in Table~\ref{table:Dailymail_CNN_all}. We can find that our approach performed competitive results compared with NN-SE \citep{cheng2016neural}  and $\rm M^{2}$SM \cite{Fu2021MMAVSAF} for the quality of generated textual summary. For visual summarization comparison, we can find that the quality of generated visual summary by our approach still outperforms the other visual summarization baselines.

\begin{table*}[htp]\small
\centering
\caption{Comparisons of multimodal baselines on the Daily Mail and CNN datasets.}
\vspace{-5pt}
\begin{adjustbox}{width=0.8\linewidth}
\begin{tabular}{l|l|ccccccc} 
\toprule
\multirow{2}{*}{Category} &\multirow{2}*{Methods} & \multicolumn{3}{c}{CNN dataset} & \multicolumn{4}{c}{Daily Mail dataset}   \\ 
\cmidrule(r){3-5}\cmidrule(r){6-9} &
&R-1 &R-2 &R-L &R-1  &R-2  &R-L &Cos(\%)   \\
\midrule
\multirow{2}{*}{Video summarization baselines}
&VSUMM   &-- &-- &-- &-- &-- &-- &68.74\\
&DR-DSN   &-- &-- &-- &-- &-- &-- &68.69\\
\midrule
\multirow{2}{*}{Textual summarization baselines}
&Lead3   &-- &-- &-- &41.07 &17.87 &30.90 &--\\
&NN-SE   &-- &-- &--  &41.22 &18.15 &31.22 &--\\
\midrule
\multirow{6}{*}{Multimodal baselines}
&VistaNet  &9.31 &3.24 &6.33 &18.62 &6.77 &13.65 &-    \\ 
&MM-ATG   &26.83 &8.11 &18.34 &35.38  &14.79  &25.41 &69.17    \\
&Img+Trans    &27.04 &8.29 &18.54 &39.28  &16.64  &28.53  &-   \\
&TFN  &27.68 &8.69 &18.71 &39.37  &16.38  &28.09 &-       \\  
&HNNattTI  &27.61 &8.74 &18.64 &39.58  &16.71 &29.04 &68.76    \\ 
&$\rm M^{2}$SM  &27.81  &8.87  &18.73 &41.73 &18.59 &31.68 &69.22     \\
\midrule
\multirow{3}{*}{Ours} 
&Ours-textual  &-- &-- &-- &40.28 &17.93 &31.89  &--\\
&Ours-video  &-- &-- &-- &-- &-- &--  &70.56   \\ 
&Ours-Multimodal   & \textbf{28.02}   & \textbf{8.94}    & \textbf{18.89}      & \textbf{44.52}  & \textbf{19.87}  & \textbf{35.79}    & \textbf{73.19}       \\ 
\bottomrule
\end{tabular}
\end{adjustbox}
\label{table:Dailymail_CNN_all}
\end{table*}

\subsection{Ablation Study}
To evaluate each component's performance, we performed ablation experiments on different modalities and different datasets. For the VMSMO dataset, we compare the performance of using only visual information, only textual information, and multimodal information. The comparison result is shown in Table~\ref{table:VMSMO_all}. We also carried out experiments on different modalities using Daily Mail  dataset to show the performance of unimodal and multimodal components, and the results are shown in Table~\ref{table:Dailymail_CNN_all}.

For the ablation results, when only textual data is available, we adopt BERT~\cite{Devlin2019BERTPO} to generate text embeddings and K-Means clustering to identify sentences closest to the centroid for textual summary selection. While if only video data is available, we solve the visual summarization task in an unsupervised manner, where we use K-Means clustering to cluster frames using image histogram and then select the best frame from clusters based on the variance of laplacian as the visual summary.

From Table~\ref{table:VMSMO_all} and Table~\ref{table:Dailymail_CNN_all}, we can find that multimodal methods outperform unimodal approaches, showing the effectiveness of exploring  the relationship and taking advantage of the cross-domain alignments of generating high-quality summaries.

\subsection{Interpretation}
\begin{figure}[htp]
  \centering
  \includegraphics[width=0.95\linewidth]{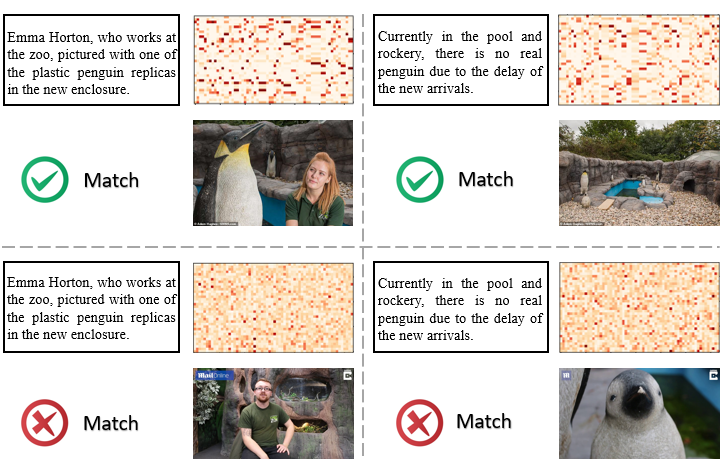}
  \vspace{-8pt}
  \caption{
  The OT coupling shows sparse patterns and specific temporal structure for the embedding vectors of groundtruth matched video and text segments.}
  \label{Fig:plan}
\end{figure}

To show a deeper understanding of the multimodal alignment between the visual domain and language domain, we compute and visualize the transport plan to provide an interpretation of the latent representations, which is shown in Figure~\ref{Fig:plan}. 
When we are regarding the extracted embedding from both text and image spaces as the distribution over their corresponding spaces, we expect the optimal transport coupling to reveal the underlying similarity and structure. Also, the coupling seeks sparsity, which further helps to explain the correspondence between the text and image data.

Figure~\ref{Fig:plan} shows comparison results of matched image-text pairs and non-matched ones. The top two pairs are shown as matched pairs, where there is overlapping  between the image and the corresponding sentence. The bottom two pairs are shown as non-matched ones, where the overlapping of meaning between the image and text is relatively small. The correlation between the image domain and the language domain can be easily interpreted by the learned transport plan matrix. 
In specific, the optimal transport coupling shows the pattern of sequentially structured knowledge. However, for non-matched image-sentences pairs, the estimated couplings are relatively dense and barely contain any informative structure. 
As shown in Figure~\ref{Fig:plan}, we can find that the transport plan learned in the cross-domain alignment module demonstrates a way to align the features from different modalities to represent the key components. The visualization of the transport plan contributes to the interpretability of the proposed model, which brings a clear understanding of the alignment module. 

\section{Conclusion}
In this work, we proposed SCCS, a segment-level Semantics-Consistent Cross-domain Summarization model for the MSMO task. Our model decomposed the video \& article into segments based on the content to preserve the structural semantics, and explored the cross-domain semantics relationship via optimal transport alignment at the segment level. The experimental results on three MSMO datasets show that SCCS outperforms previous summarization methods. We further provide interpretation by the OT coupling. Our approach provides a new direction for the MSMO task, which can be extended to many real-world applications.

\bibliography{anthology,custom}
\bibliographystyle{acl_natbib}

\end{document}